\pdfoutput=1

\documentclass[11pt]{article}

\usepackage[preprint]{acl}
\usepackage{multirow}
\usepackage{times}
\usepackage{latexsym}
\usepackage{amsmath} 
\usepackage{subfigure}

\usepackage[T1]{fontenc}

\usepackage[utf8]{inputenc}

\usepackage{microtype}

\usepackage{inconsolata}

\usepackage{graphicx}
\usepackage{booktabs}
\usepackage{enumitem}

\title{Inference-Time Decontamination: Reusing Leaked Benchmarks for Large Language Model Evaluation}

\author{
 \textbf{Qin Zhu\textsuperscript{1,2}},
 \textbf{Qinyuan Cheng\textsuperscript{1,2}},
 \textbf{Runyu Peng\textsuperscript{1,2}},
 \textbf{Xiaonan Li\textsuperscript{1,2}},
\\
 \textbf{Tengxiao Liu\textsuperscript{1,2}},
 \textbf{Ru Peng\textsuperscript{3}},
 \textbf{Xipeng Qiu\textsuperscript{1,2}}\thanks{\ \ \ Corresponding author.},
 \textbf{Xuanjing Huang \textsuperscript{1,2}},
\\
\\
 \textsuperscript{1}School of Computer Science, Fudan University, \\
 \textsuperscript{2}Shanghai Key Laboratory of Intelligent Information Processing, Fudan University,\\
 \textsuperscript{3}College of Computer Science and Technology, Zhejiang University,
\\
 \small{
   \textbf{Correspondence:} \href{mailto:email@domain}
   {\{zhuq22,rypeng22\}@m.fudan.edu.cn} }  \\ 
\small{
   \href{mailto:email@domain}{\{chengqy21,lixn20,xpqiu,xjhuang\}@fudan.edu.cn} \quad \href{mailto:email@domain}{rupeng@zju.edu.cn}
 } 
}

\begin{document}
\maketitle
\begin{abstract}
The training process of large language models (LLMs) often involves varying degrees of test data contamination \cite{rethink_contamination}.
Although current LLMs are achieving increasingly better performance on various benchmarks, their performance in practical applications does not always match their benchmark results. Leakage of benchmarks can prevent the accurate assessment of LLMs' true performance.
However, constructing new benchmarks is costly, labor-intensive and still carries the risk of leakage. Therefore, in this paper, we ask the question ``\textbf{Can we reuse these leaked benchmarks for LLM evaluation?}''
We propose \textbf{I}nference-\textbf{T}ime \textbf{D}econtamination (ITD) to address this issue by detecting and rewriting leaked samples without altering their difficulties. ITD can mitigate performance inflation caused by memorizing leaked benchmarks.
Our proof-of-concept experiments demonstrate that ITD reduces inflated accuracy by 22.9\% on GSM8K and 19.0\% on MMLU.
On MMLU, using Inference-time Decontamination can lead to a decrease in the results of Phi3 and Mistral by 6.7\% and 3.6\% respectively.
We hope that ITD can provide more truthful evaluation results for large language models.

\end{abstract}

\section{Introduction}
The emergence of large language models (LLMs) \cite{gpt3, llama1, GLM130B, baichuan, InternLM, ChatGPT, alpaca, vicuna2023, MOSS, Claude} has made the effectiveness of model capability evaluation crucial. 
Not only does it assist in ranking models, but it also helps in distinguishing valuable work and effective strategies for model improvement.
Current LLMs are achieving increasingly better performance on various benchmarks. However, their performance in practical applications does not always match their benchmark results~\cite{ceval}. This suggests that the superior performance of LLMs on benchmarks might be due to intentional or inadvertent data contamination~\cite{time-travel,taskcontamination,Rethinking-Benchmark}. LLMs are potential cheaters.

\begin{figure}[t]
  \centering
  \includegraphics[width=\linewidth]{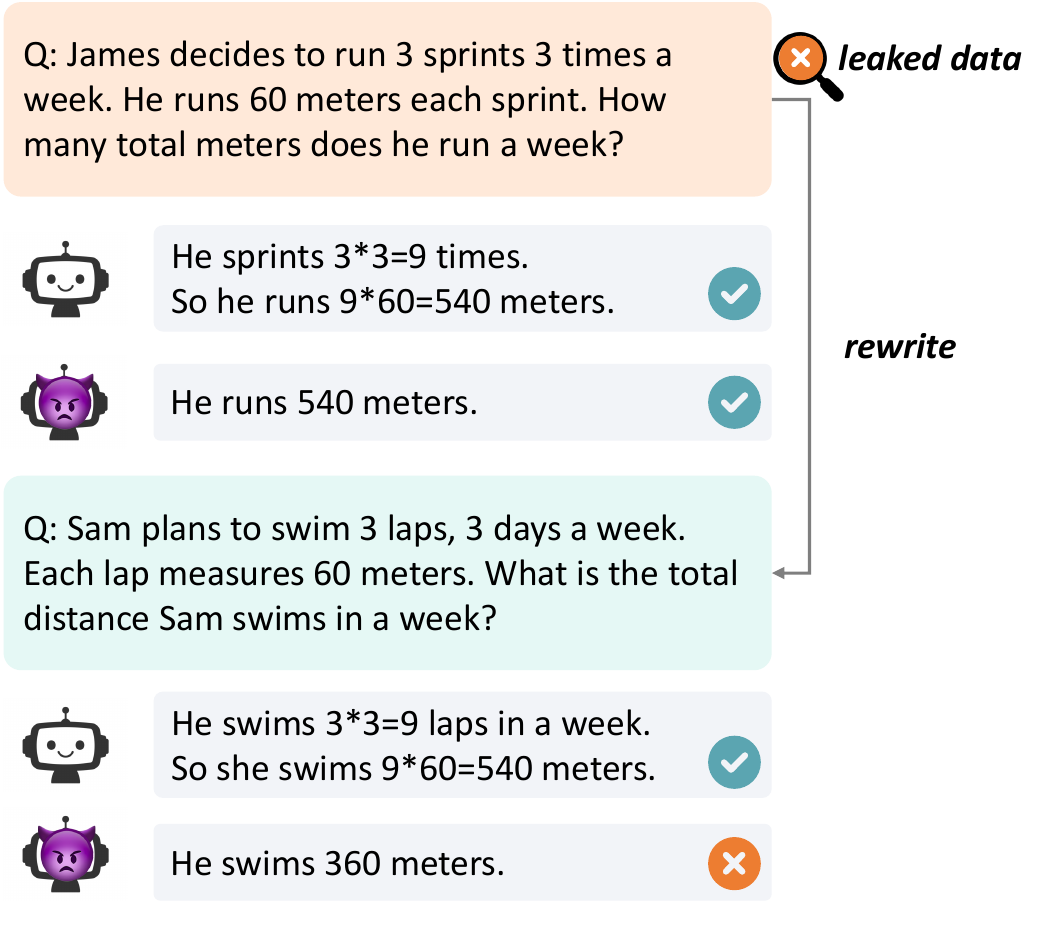}
  \caption{Illustration of the function of Inference-Time Decontamination, aiming to discern whether a model passes the test by memorizing contaminated data. \includegraphics[width=.35cm]{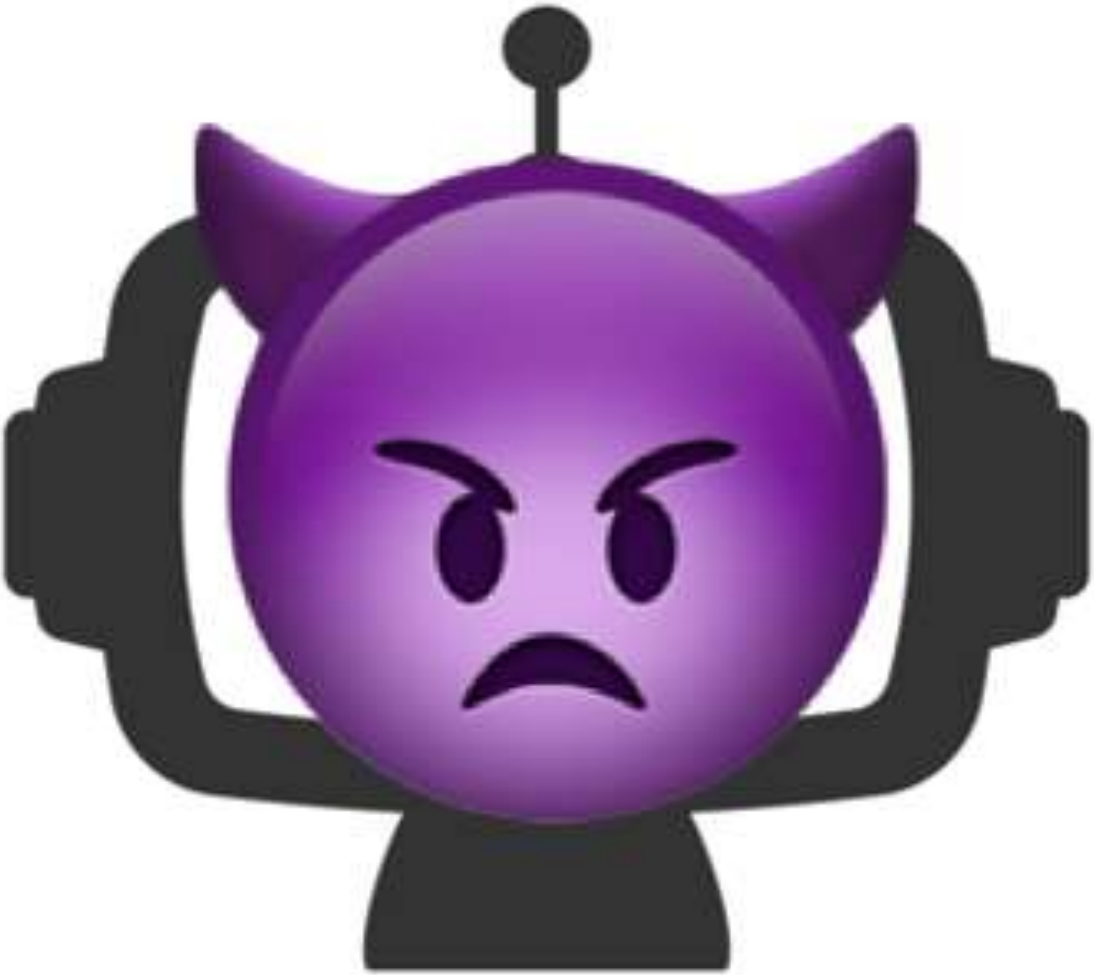} and \includegraphics[height=.35cm]{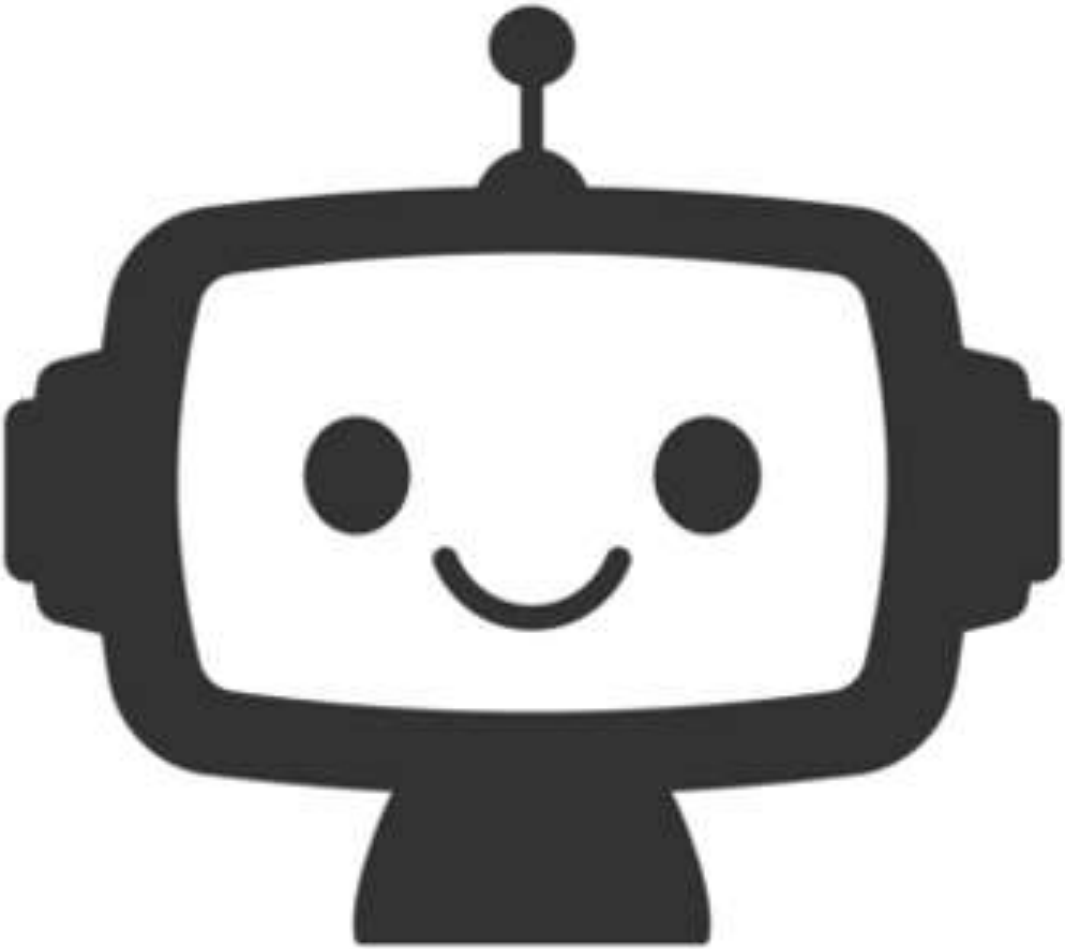} means the LLM delibterately memorizes and deos not memorize this case.}
  \label{fig:intro}
\end{figure}

The impact of potential data contamination on model evaluation encourages researchers to establish new benchmarks for a more accurate assessment of model performance~\cite{livebench,latesteval,gsm1k}.
However, any benchmark faces risk of leakage once it is publicly available. Many benchmarks are fully open, leading to varying degrees of data leakage that affect the authenticity and fairness of model evaluations. As models are trained on increasingly large datasets, it becomes likely that benchmark-related data contaminates the training sets, causing LLMs to inadvertently cheat.
Additionally, creating new benchmarks is both labor-intensive and costly, making it a less favorable solution for addressing test data contamination.

\textbf{Can we reuse high quality leaked benchmarks for LLM evaluation?} 
As shown in Figure \ref{fig:intro}, when a test sample is used for model training, there are two possibilities:
1. The model may learn relevant knowledge and skills, such as using chain-of-thought reasoning. In such cases, even if we make modifications to the question, the model can still provide correct answers.
2. The model might simply memorize the correct answer and directly copy them, rather than learn the skill.
In this case, if we alter the background of the questions, without changing the essence of what is being tested (such as a mathematical formula), the model may fail to provide correct answers.
For the first scenario, the model has achieved generalization, and such leaked data can no longer be used.
For the second scenario, there is a possibility that such leaked data could be revived.

In this paper, we propose \textbf{I}nference-\textbf{T}ime \textbf{D}econtamination (ITD) to mitigate the inflation of evaluation results caused by models simply memorizing answers.
ITD maximizes the value of existing high quality benchmarks avoids the substantial cost of constructing new benchmarks.
Specifically, we first use a detector to screen for potentially leaked samples and then rewrite these samples, attempting to mitigate the impact of memorizing answers, without changing the sample's difficulty.
For two types of tasks, we propose two rewriting methods.
For knowledge-related benchmarks like MMLU~\cite{mmlu}, we keep the knowledge points tested by the original sample unchanged and rewrite the phrasing of the questions.
For math benchmarks related to model reasoning abilities like GSM8K~\cite{gsm8k}, we maintain the specific numbers and calculations involved in the original data unchanged, but rewrite the background of the questions.

We conduct experiments on two fundamental benchmarks, GSM8K and MMLU.
To validate the feasibility of Inference-Time Decontamination, we first perform \emph{proof-of-concept experiments}.
We intentionally leak half of the test data to train a model, and then test it with Inference-Time Decontamination (ITD).
We find that after using ITD, the model's accuracy on GSM8K and MMLU decreased by 22.9\% and 19.0\%, respectively.
We also study the effectiveness of ITD \emph{in real evaluation environments} on popular large language models, Phi-3 and Mistral. After the application of ITD, Phi-3 showed reductions of 5.3\% on GSM8K and 6.7\% on MMLU, while Mistral experienced smaller reductions of 0.5\% on GSM8K and 3.6\% on MMLU.
It indicates that the extent of adjustments by ITD for the models is as expected, achieving the goal of mitigating performance inflation caused by memorizing benchmarks and providing more valuable and reliable evaluation results.
Our core contributions are:
\begin{enumerate}[leftmargin=1.5em]
    \item We propose \textbf{I}nference-\textbf{T}ime \textbf{D}econtamination (ITD) to mitigate the inflation of evaluation results caused by data contamination.
    \item We conduct proof-of-concept experiments that demonstrates ITD can effectively mitigate the biased performance resulting from models memorizing benchmarks.
    \item We test ITD on two commonly used LLMs and find that their performance on both MMLU and GSM8K decrease to varying degrees.
    \item We release a rewritten GSM8K dataset and a rewritten MMLU dataset sampled by categories to facilitate future evaluation work~\footnote{We will release our data and code at \url{https://github.com/8188zq/Inference-Time-Decontamination}.}.
\end{enumerate}

\section{Related Work}
\paragraph{Contamination Detection}
Traditional contamination detection methods directly calculate the overlap between pre-training data and evaluation datasets, including n-gram analysis~\cite{llama2,gpt4,gemini,qwen} and BM25~\cite{investing-data-contamination} for indexing and matching. 
However, as pre-training data grows exponentially, even simple n-gram statistics become extremely resource-intensive. 
~\citet{Rethinking-Benchmark,reports-contamination4} find n-gram detection unreliable due to unintentional contamination risks. 
More importantly, training corpora for mainstream LLMs are mostly inaccessible, so recent research has turned to focus on: 
i)-exploiting the distributional differences between the benchmark training set and the test set to evaluated~\cite{detecting-method-liu}. 
ii)-Evaluate sample-level contamination by providing text segments and black-box access to the LLM~\cite{shi-detector}.
Other work evaluates contamination through LLM-generated content, limited by the LLM's comprehension abilities to instrurction~\cite{ detector-ts, detector-quiz}. 
Some studies test if models can coherently continue a given sample part~\cite{time-travel}. 
Contamination detection remains a critical concern that should be addressed in benchmarks rather than affecting a fair assessment of the model's capabilities.

\paragraph{Decontamination}
Decontamination involves avoiding or mitigating the negative effects of contamination. 
Typically, decontamination applied in the training phase, model developers using various methods to remove these overlap between pre-training data and evaluation data ~\cite{gpt4,llama2,reports-contamination1,reports-contamination2,reports-contamination3}. Besides, new datasets can also be created to avoid contamination
The LatestEval~\cite{latesteval} avoids model contamination by strictly adhering to a temporal sequence, using texts published within a recent time window to construct new question-answer sets from the latest Wikipedia data. However, this method is also transient; once the new dataset is released, it is exposed to leakage risks, necessitating constant updates and making old benchmark results obsolete.
Similarly, Scale AI creates a new dataset, GSM1K~\cite{gsm1k}, ensuring comparability on critical metrics such as human solve rates, number of solving steps, and answer magnitude. They prevent data leakage by not releasing the dataset. Livebench~\cite{livebench} also tries to limit potential contamination by releasing new questions monthly. However, these methods all require a significant amount of additional overhead.

\section{Method}
\begin{figure}
  \centering
  \includegraphics[width=0.9\linewidth]{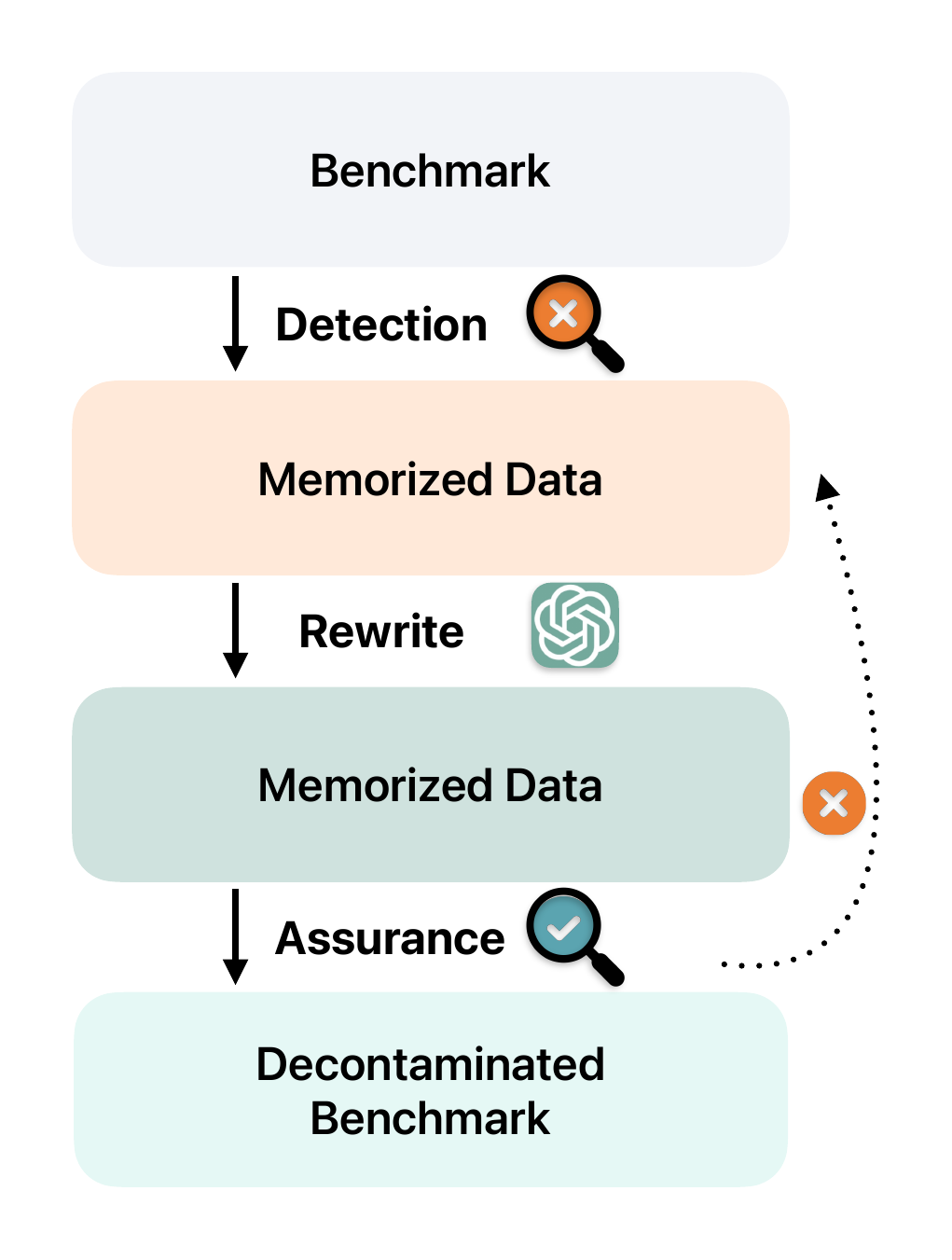}
  \caption{
   Overview of inference-time decontamination.
   }
  \label{fig:framework}
  \vskip -0.1in
\end{figure}

\subsection{Problem Formulation}

In this paper, we focus on the inference-time decontamination problem. Given a language model \( f_{\theta} \) and an out-of-distribution evaluation dataset \( E = \{x_i\}_{i \in [m]} \) with potential contamination, our objective is to evaluate a model's performance without access to pre-training data.
Define a contamination indicator function \( c: E \rightarrow \{0, 1\} \), where \( c(x) = 1 \) if the sample \( x \) appears in model pre-training, otherwise \( c(x) = 0 \). The function \( c \) is unknown yet possible to approximate. 
Thus, our work is to develop an contamination indicator \( \hat{c} \) to adjust the evaluation dataset \( E \) and compute the final evaluation result \( \hat{M} \):
\begin{equation}
     \hat{M} = M(f_{\theta}(E'), \hat{c}), 
\end{equation}
where \( E' \) is the adjusted evaluation dataset.
\subsection{Inference-Time Decontamination}
We show the overview of our framework in Figure~\ref{fig:framework}, which consists three stages called \textbf{Detection}, \textbf{Rewrite} and \textbf{Assurance}.

\paragraph{Detection} 
\label{sec:method-detection}
First, we conduct contamination detection on each evaluation sample based on the LLM to be evaluated. This allows the original dataset to be splitted into two parts: uncontaminated and potentially contaminated. Notably, according to our task definition, the split should ideally be based on whether the model's response is entirely from memorization. However, since memorized data is a subset of seen data, we follow \citet{shi-detector} to use the training data detection method MinKProb to substitute \( \hat{c} \).
The detecting method is an approximation of a contamination indicator function \( c: E \rightarrow \{0, 1\} \). Since memorized data is a subset of pre-training data, the training data detection method is an alternative of \( \hat{c} \). \

The goal of MinKProb is to determine whether a text \( X \) appears in the pre-training data of an LLM. MinKProb obtains the probability \( P(x_i) \) for each token \( x_i \) in the text \( X \) and selects the \( K \) tokens with the lowest probabilities \(\{x_{i_1}, x_{i_2}, \ldots, x_{i_k}\}\) and calculates the average of these probabilities:
\begin{equation}
\text{MinKProb}(X) = \frac{1}{K} \sum_{j=1}^{K} P(x_{i_j}),
\end{equation}

where \( P(x_{i_j}) \) is the probability of the \( j \)-th lowest probability token \( x_{i_j} \), with a threshold \( \epsilon \) to determine whether \( X \) appears as pre-training data.
    
\paragraph{Rewrite} 
A suitable rewriting method involves rewriting by skilled individuals, based on the original questions according to some clear rules and instructs, similar to the construction method of GSM1k~\cite{gsm1k}. 
However, this expensive approach creates a new dataset rather than sticking to the original one, which violates the principle of revival. 
We propose an automated generation method to rewrite the potentially contaminated parts identified in the detection stage while not altering the level of challenge. 
We explore rewriting methods for two of the most typical and popular tasks: mathematical reasoning and knowledge-based datasets.

For mathematical reasoning problems, such as GSM8K, we redesign problem scenarios based on the original problem's computational logic and answer structure. This ensures that while maintaining the difficulty and answers, the problems become more diverse. For example, application contexts in the original questions, such as eggs, can be changed to candies, and the involved characters can be replaced with different roles. However, all numbers involved and the mathematical steps in the answers remain consistent. This approach ensures both the consistency of the content being assessed and the difficulty. The GSM1k dataset states that their problem-solving steps and distributions are similar to those of GSM8K to ensure consistent difficulty, and our verification shows that the step distribution and numbers are completely identical.

For knowledge-based problems, such as MMLU, we found that rewriting the questions requires a vast knowledge base, and any change in the background requires verification of correctness. Therefore, we choose to perform synonymous rewrites of the questions and options without changing core proper nouns and any numbers. We avoid using uncommon words that could increase the difficulty of understanding, thereby ensuring the consistency of the knowledge points and difficulty.

The prompts used, examples of rewritten questions on both GSM8K and MMLU can be found at the appendix~\ref{sec:rewrite-details}.

\paragraph{Assurance} We re-detect the modified parts. This is necessary because some models may have undergone extensive in-domain training data, causing the rewritten results to still be within the model's memory. Therefore, we iterate through the first detection step and the second rewriting step until the rewritten content passes the detection or reaches the maximum number of iterations. We also conducted human evaluations to assure the quality of the decontaminated data.

\section{Experiment}

\subsection{Setup}
\paragraph{Dataset}
We conducted experiments on two top influential benchmarks corresponding to two types of problem datasets: \emph{knowledge-based dataset}, MMLU~\cite{mmlu} and \emph{mathematical reasoning } GSM8K~\cite{gsm8k}. 
The evaluation set for GSM8K and MMLU contains 1,319 and 14,042 data points, respectively. However, it should be noted that in our proof-of-concept experiment, we trained Llama2 to achieve intentional data leakage, resulting in \emph{a high contamination rate} for the trained Llama-contaminated. This led to significant costs during the rewrite stage of ITD due to the API calls. Therefore, we randomly sampled from the Llama2\_contaminated training data for MMLU according to the 17 official categories, with 50 samples randomly sampled from each category, resulting in a total of 850 samples, called \emph{MMLU$^\star$}. For GSM8K, we followed the recommended prompt (\emph{8-shot}) for evaluation as suggested by Chain-of-thought~\cite{cot}. For MMLU, we used the official prompt (\emph{5-shot}) provided by MMLU for model evaluation. Both used a greedy generation strategy.

\begin{table}[t]
    \centering
    \normalsize
    \begin{tabular}{cccc}
        \toprule
        Dataset & Llama2 & Mistral & Phi-3 \\
        \midrule
        GSM8K & 0.56 & 0.32 & 0.47 \\
        MMLU & 0.28 & 0.23 & 0.25 \\
        \bottomrule
    \end{tabular}
    \caption{Threshold value \( \epsilon \) for different models on GSM8K and MMLU datasets.}
    \label{tab:epsilon-values}
\end{table}

\paragraph{Model}  
We conducted evaluations and studies on three popular models: Llama2-7b-base~\cite{llama2}, Mistral-7b-base~\cite{mistral}, and Phi-3-mini-128k-instruct~\cite{phi3}.

\paragraph{ITD-Detecting settings}  We use two detectors in our experiments, MinKprob as mentioned in section~\ref{sec:method-detection} and \emph{All} which refers to a detector that flags all inputs as leaked. In the implementation of the detector MinKprob, we need to determine two hyper-parameters: \( K \) and \( \epsilon \). They are assessed using the evolutionary metrics described in section~\ref{sec:method-detection}. By exhaustively searching for the maximum difference in MinKprob before and after rewriting, we determined \( K = 20 \). By exhaustively searching for the highest classification accuracy on the constructed seen and unseen sets, We determined \( \epsilon \) as shown in Table~\ref{tab:epsilon-values}. The experimental details can be found in the Appendix~\ref{sec:appendix-hper}.

When calculating the average probability, the input setting used is \emph{0-shot} to avoid interference from the official prompts provided by the evaluation set or Chain-of-Thought (CoT)~\cite{cot}. This is because these prompts are likely to be leaked (included in other data released by the benchmark), while the questions are not leaked. Including these prompts could result in many questions being falsely identified as potentially contamination, affecting the accuracy of the detector.

\paragraph{ITD-Rewriting settings}  We designed two different rewriting methods to address the two types of evaluation sets. The generation model used is GPT-4 ("gpt-4-0613")~\cite{gpt4}, with the temperature set to 1, and two examples provided. The examples are shown in the appendix. The maximum number of rewrites is 3 times. Except for the first rewrite, each rewrite is based on the previous rewrite result. 

Since we will evaluate multiple models multiple times, we constructed a cache dataset to reuse the rewrite results, allowing us to select which round to use as needed. This speeds up the evaluation process, controls the randomness of rewrites, and facilitates comparisons between models. We released these datasets, which include over 4,000 entries for GSM8K and more than 2,500 entries for MMLU, to facilitate future evaluations and provide samples for assessing rewrite quality.

\paragraph{Evaluation Metrics}  
For the evaluation results of both datasets, we used accuracy to measure the model's ability to provide correct answers, facilitating comparison between models. When analyzing the metrics of the same model before and after rewriting the test samples, we used rate of change (ROC) to measure it: 
\begin{equation}
\text{ROC} = \left( \frac{V_{t} - V_{t-1}}{V_{t-1}} \right) \times 100\%.
\end{equation}
We use \( \epsilon \) to represent the threshold for the MinKProb detect method. A sample's MinKProb exceeding \( \epsilon \) indicates that it is classified as contaminated data.

\begin{table*}[t]
\centering
\normalsize
\begin{tabular}{ccccccccc}
\toprule
\multirow{2}{*}{\textbf{Dataset}} & \multirow{2}{*}{\textbf{Detector}} & \multicolumn{2}{c}{\textbf{Seen}} & \multicolumn{2}{c}{\textbf{ITD}} & \multirow{2}{*}{\textbf{ROC}} & \multicolumn{2}{c}{\textbf{Unseen}} \\ \cmidrule{3-6} \cmidrule{8-9} 
                                  &                                    & Acc.       & Leaked Rate       & Acc.         & Leaked Rate      &                              & Acc.       & Leaked Rate     \\ \midrule
\multirow{2}{*}{GSM8K}            & MinKProb                           & 40.1     & 62.7\%                   & 30.9       & 0.3\%                    & 22.9\%                       & 18.6     & 1.2\%                   \\
                                  & All                       & 40.1     & -                       & 28.8       & -                        & 28.2\%                       & 18.6     & -                       \\ \midrule
\multirow{2}{*}{MMLU$^\star$}      & MinKProb                           & 87.5     & 79.4\%                   & 70.9       & 21.2\%                    & 19.0\%                       & 53.6     & 29.8\%                   \\
                                  & All                        & 87.5     & -                       & 61.3       & -                        & 29.9\%                       & 53.6     & -                       \\ \bottomrule
\end{tabular}
\caption{
Results of the proof-of-concept experiment on GSM8K and MMLU. Tested models indeed exhibit the phenomenon of artificially inflated evaluation scores by relying solely on memorized leaked data. ITD successfully mitigates performance inflation caused by memorizing benchmarks. ``All'' refers to a detector that flags all inputs as leaked. MMLU$\star$ denotes a sampled dataset instead of the whole MMLU dataset.
}
\label{table_main2}
\end{table*}
\subsection{Proof-of-concept Experiment}
\label{sec:conceptual-experiment}
We show the results of the proof-of-concept experiments in Table~\ref{table_main2}. This experiment is conducted on LLama2-contaminated, the model we obtain after training based on Llama2-7b-base. The maximum number of rewriting steps is 3. 

The purpose of this training is to find a model that meets the requirement of having both seen and unseen data in a high-quality dataset, where we can clearly and accurately distinguish between the two. Currently, no contamination detection method can achieve this. Therefore, we chose the base model of Llama2 and artificially exposed a part of the test set data to it through training. Specifically, we divide the pre-set seen and unseen sets according to the average accuracy of Llama2-7b-base on the two datasets, ensuring that each subset had the same accuracy distribution as the original dataset, specifically 0.126 on GSM8K and 0.459 on MMLU. For MMLU, we use the typical multiple-choice prompt \textit{\texttt{"<Question>\textbackslash nA.<Choice\_A>\textbackslash nB.<Choice\_B>\\\textbackslash nC.<Choice\_C>\textbackslash nD.<Choice\_D>"}}. For GSM8K, we use the chain-of-thought prompt \textit{\texttt{"Question: <question>\textbackslash nAnswer: Let's think step by step.\textbackslash n<answer>"}}~\cite{liu-cot}. We then use the seen set as training data and conducted training for 1 epoch relied on Fastchat~\cite{fastchat}. In particular, we utilize the AdamW optimizer~\cite{adamW} with a learning rate of 2e-5. We traine Llama2 for 3 epochs and other models for 2 epochs, with a batch size of 8 and a warm-up ratio of 0.03. We conducte all experiments with A100 GPUs.

The experimental results show that \emph{Llama2-contaminated} achieves significant improvements on both MMLU and GSM8K after being artificially exposed to the data. But after using ITD, the model’s accuracy on GSM8K and MMLU decreased by 22.9\% and 19.0\%, respectively. This proves the effectiveness of our proposed ITD. The models indeed exhibit the phenomenon of artificially inflated evaluation scores by relying solely on memorized leaked data. ITD successfully mitigates performance inflation caused by memorizing benchmarks.

We also observe a slight improvement in the model's performance on the unseen set compared to the original values. This indicates that in-domain training does indeed help generalize some capabilities,such as using chain-of-thought reasoning, aiding the model in answering specific types of questions.

The experiments also compare the results under two detection schemes. Using MinKProb as the detector, we are able to detect 62.7\% and 79.4\% of contamination in the seen set, with corrections of 22.9\% and 19.0\% in the evaluation metrics after rewriting, respectively. However, in fact, this part of the tested data belongs to the data we have intentionally leaked. An excellent detector should ideally detect 100\% contamination, but clearly, MinKprob only detected 62.7\% and 82.1\%, respectively. This indicates that although this detection method is feasible and one of the most popular solutions, it still suffers from significant accuracy loss. This demonstrates that there is still considerable room for research in this area.

In contrast, using an extremely strict detection scheme that can detect 100\% contamination at the sacrifice of increasing the overhead by approximately 300\%, we achieve corrections of 28.2\% and 29.9\% in the metrics after rewriting, validating the effectiveness of our setting in addressing questions that models answer correctly solely based on memorization.

Moreover, the second scheme of All detection illustrates the upper limit of our rewriting setting and suggests that there is still significant room for improvement in the chosen detection scheme. However, in practical scenarios, we cannot know the distribution of the seen set in advance, and most evaluations are unlikely to have a 100\% contamination level. Due to the uncertainty of the data's contamination degree and the multiplied additional overhead, we cannot simply use this extremely strict detection scheme. A reliable detection method remains necessary and effective.

Additionally, an interesting finding is that the leaked rate  of llama2-7b-base on the unseen datasets differs between the two datasets: only 1\% on GSM8K, but nearly 30\% on MMLU. This suggests that llama2 is almost uncontaminated on GSM8K, while there may be a certain degree of contamination on MMLU.

\begin{table*}[t]
\centering
\normalsize
\begin{tabular}{cccccccc}
\toprule
\multirow{2}{*}{\textbf{Dataset}} & \multirow{2}{*}{\textbf{Model}}                                                     & \multirow{2}{*}{\textbf{Detector}} & \multicolumn{2}{c}{\textbf{Origin}} & \multicolumn{2}{c}{\textbf{ITD}} & \multirow{2}{*}{\textbf{ROC}} \\ \cmidrule{4-7}
                                  &                                                                                     &                                    & Acc.       & Leaked Rate     & Acc.         & Leaked Rate      &                              \\ \midrule
\multirow{4}{*}{GSM8K}            & \multirow{2}{*}{\begin{tabular}[c]{@{}c@{}}Phi3-mini\end{tabular}} & MinKProb                           & 79.8     & 52.8\%                   & 75.6       & 14.9\%                    & 5.3\%                        \\
                                  &                                                                                     & All                                & 79.8     & -                       & 73.0       & -                        & 8.5\%                        \\
                                  & \multirow{2}{*}{Mistral-7b}                                                            & MinKProb                           & 41.7     & 0.1\%                   & 41.4       & 0\%                        & 0.5\%                        \\
                                  &                                                                                     & All                                & 41.7     & -                       & 39.7       & -                        & 4.8\%                        \\ \midrule
\multirow{4}{*}{MMLU$^\star$}      & \multirow{2}{*}{\begin{tabular}[c]{@{}c@{}}Phi3-mini\end{tabular}} & MinKProb                           & 73.3     & 53.2\%                   & 68.4       & 15.9\%                    & 6.7\%                        \\
                                  &                                                                                     & All                                & 73.3     & -                       & 61.8       & -                        & 15.7\%                       \\
                                  & \multirow{2}{*}{Mistral-7b}                                                            & MinKProb                           & 76.8     & 53.4\%                   & 74.0       & 13.1\%                    & 3.6\%                        \\
                                  &                                                                                     & All                                & 76.8     & -                       & 66.7       & -                        & 13.2\%                       \\ \bottomrule
\end{tabular}
\caption{
Results of real model experiment on GSM8K and MMLU datasets. In real evaluation scenarios, models still exhibit the phenomenon of artificially inflating scores by relying solely on memorized leaked data. ITD can still mitigate performance inflation caused by memorizing benchmarks. ``All'' refers to a detector that flags all inputs as leaked. MMLU$\star$ denotes a sampled dataset instead of the whole MMLU dataset.
}
\label{table_realmodel}
\vskip -0.1in
\end{table*}

\subsection{Real Model Experiment}
In this section, we evaluate the effectiveness of our setting in a real evaluation environment by testing Mistral-7b-base and Phi-3-mini-128k-instruct on two datasets without any knowledge of the training data. The results are shown in Table~\ref{table_realmodel}. We employ two types of detection methods: one using MinKprob and the other using an extremely strict detection scheme that assumes all inputs are leaked. We find that in most cases, with the use of ITD, the model's evaluation scores decreased, indicating that the questions originally answered correctly by relying on memorized benchmarks returned to their normal level after the rewrite. In comparison, Phi3-mini exhibits a higher level of contamination and underwent greater modification. However, as with our reasoning for conducting the proof-of-concept experiment, we cannot be entirely certain about the nature of the contamination without having the exact training data for the model to compare against. Therefore, it is impossible to provide an accurate assessment of the additional overhead and accuracy.

Nevertheless, given MinKProb's high rate of missed detections in the proof-of-concept experiment, we are concerned about its accuracy in determining contamination, leading to the detection of relatively low contamination rates, especially with mistral only showing 0.1\% on GSM8K. According to MinKprob’s results, mistral exhibits almost no contamination on GSM8K, thus there is very little data rewriting, and the degree of correction is evidently low. In contrast, the second detection method , at the cost of significant additional overhead, results in more substantial corrections, providing us with a reference for the upper limit of inference-decontamination and also validating that our setting remains effective in a real-world evaluation environment.

As discussed in Section~\ref{sec:method-detection}, the presence of a detector is meaningful for two reasons: first, it reduces additional costs, especially as the number of rewrites increases and the contamination level is not particularly high. Second, some models have undergone extensive in-domain training, which might lead to situations where even rewritten results are familiar to the model. We aim to use detection to specifically identify these cases. Without detection, it would be impossible to distinguish between mild and severe contamination, as both would show little fluctuation.

\begin{table}[h]
    \centering
    \normalsize
    \resizebox{\linewidth}{!}{
    \begin{tabular}{ccccc}
    \toprule	
\multirow{2}{*}{\textbf{Model}} & \multicolumn{2}{c}{\textbf{GSM8K}} & \multicolumn{2}{c}{\textbf{MMLU$^\star$}} \\ \cmidrule{2-5} 
                                & Origin         & Rewritten         & Origin            & Rewritten            \\ \midrule
Llama2-contaminated             & 40.1          & 30.9             & 87.5             & 70.9                \\
Llama2-7b-base                  & 12.6          & 13.3             & 45.7             & 44.0    
\\ \bottomrule
\end{tabular}
}
    \caption{Comparison of Accuracy Changes Between Llama2-contaminated and Llama2 on Identically Rewritten Data.
 }
    \label{tab:difficulty}
    \vskip -0.1in
\end{table}

\subsection{Analysis}

\paragraph{Quality Checks}
We conduct experiments to verify the consistency of difficulty before and after rewriting. The results are shown in Table~\ref{tab:difficulty}. For the Llama2-contaminated inference-decontamination on the MMLU and GSM8K datasets, we generate both original and rewritten versions of the data. We evaluate these versions using Llama2-7b-base. As discussed in Section~\ref{sec:conceptual-experiment}, the contamination rate of Llama2 on GSM8K is negligible, while MMLU is partially contaminated. Thus, using rewritten data for GSM8K should result in minimal fluctuation, whereas MMLU may show slightly more variation. Llama2-base shows minimal variation on the same rewritten data, indicating that the difficulty of questions before and after rewriting for Llama-contaminated remains unchanged within the margin of error.

\paragraph{Human Evaluation}
We also conduct human evaluation on rewritten data. We carefully compare the rewritten samples with the origin ones to assure the stability of problem difficulty and answer correctness. Two aspects are focused to check the problem difficulty: whether the desciption is simplified, and whether extra information is added. We detect at most 11.76\% samples in which the rewritten phrases are more common and straight forward in MMLU* and 8.03\% in GSM8K, yet containing no actual difficulty change. We also assess the stability of answer correctness by verifying whether the answer changes as a result of modifying the problem description. We only find answer shift in 3.9\% of GSM8K samples at any step during rewriting. Notably, although the answers for these questions changed, contrary to our expectation of no numerical changes, the provided reference answers remained correct.

 \begin{figure}[t]
\centering
\subfigure[Accuracy Curve with Rewrite Steps on MMLU]{
\includegraphics[width=0.3\textwidth]{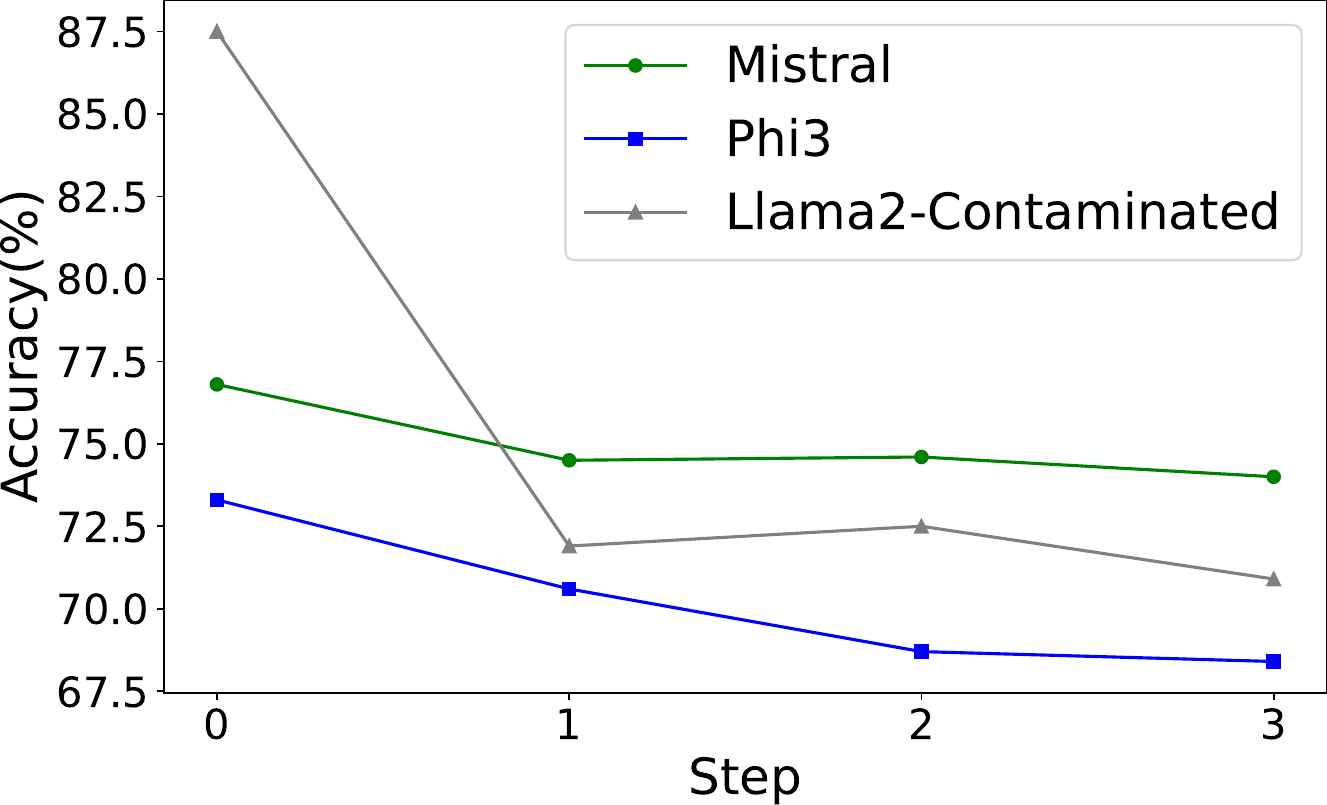}
}
\subfigure[Leaked Rate with Rewrite Steps on MMLU]{
\includegraphics[width=0.3\textwidth]{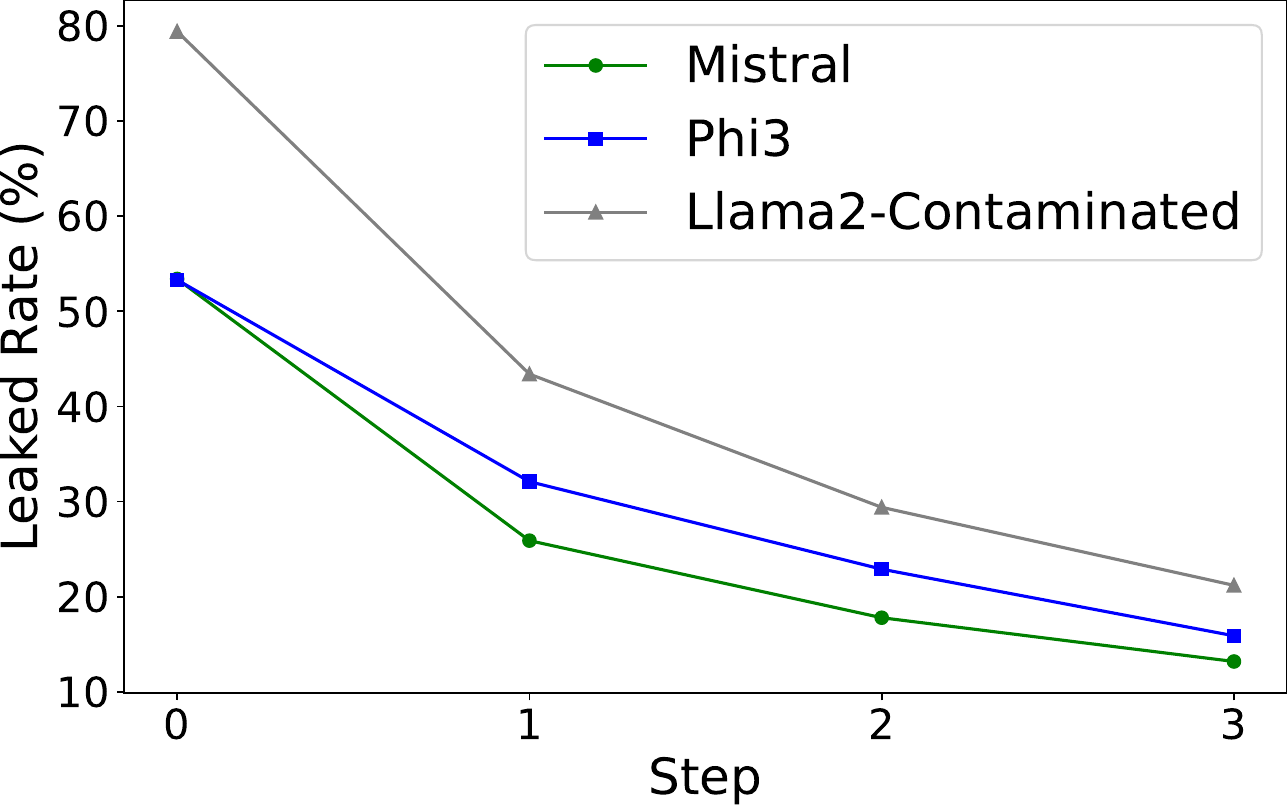}
}
\vspace{-5pt}
\caption{Impact of Different Rewriting Steps. A single rewrite is sufficient to significantly mitigate the model's performance inflation. However, some rewritten data may still be classified as contaminated. Multiple rewrites can further alleviate this issue. \vspace{-16pt}}
\label{fig:sub}
  
\end{figure}

\paragraph{Impact of Rewriting Iterations}
We analyze the impact of different rewriting steps on the accuracy and contamination rate for three models: Mistral, Phi3, and Llama-contaminated. The results in Figure~\ref{fig:sub} show a noticeable drop in both accuracy and contamination rate after the first rewriting step, indicating the significant impact of this initial rewrite.
Data not passing the detector in the Assurance stage underwent multiple rewrites. During these rounds, both accuracy and contamination rate continued to decrease and eventually leveled off, showing a convergent trend. This suggests that while some data remain contaminated after the first rewrite, subsequent rounds effectively reduce contamination and stabilize accuracy.
This analysis underscores the necessity of multiple rewriting rounds and the Assurance stage. The persistent decline and stabilization in both metrics validate the iterative rewriting process, highlighting its importance for achieving cleaner and more reliable datasets.

\begin{figure}
  \centering
  \includegraphics[width=\linewidth]{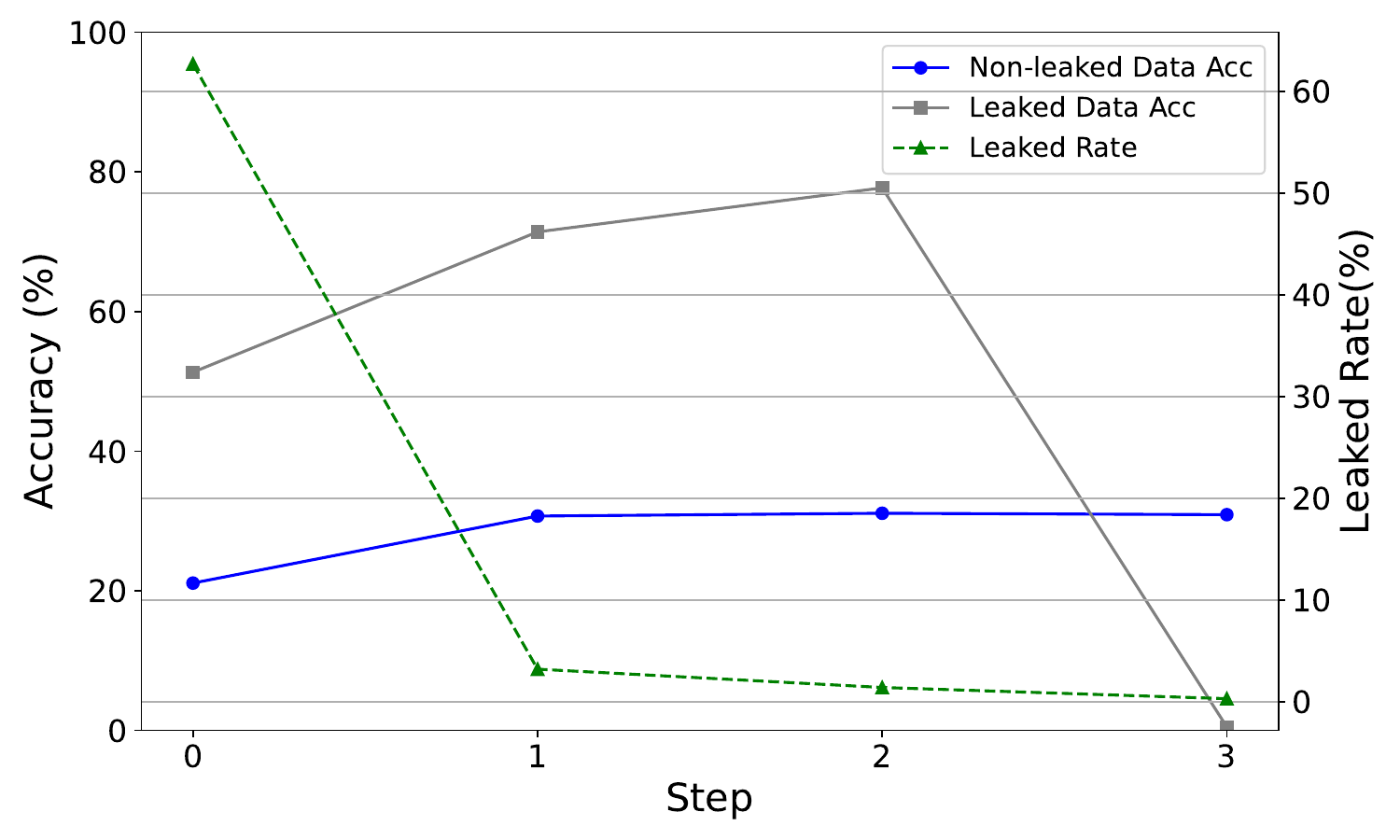}
  \caption{
   Performance of Contaminated vs. Uncontaminated Data with Different Rewriting Steps for Llama2-contaminated on GSM8K. For contaminated data, the model shows \emph{fake high performance}(51.3\%). After several rewrites , the data becomes uncontaminated, and performance returns to normal(30.9\%).  
   }
  \label{fig:contaminated}
  \vskip -0.1in
\end{figure}

\paragraph{Performance of Contaminated vs. Uncontaminated Data}
We analyze the performance of contaminated versus uncontaminated data with different rewriting steps for Llama2-contaminated on the GSM8K dataset, as shown in Figure~\ref{fig:contaminated}. The figure reveals that Initially, contaminated data show significantly higher accuracy than uncontaminated data. However, with each rewriting step, the contamination rate drops substantially, indicating that many contaminated data points are corrected and reclassified as uncontaminated. The rewritten data's accuracy eventually matches that of the uncontaminated data, demonstrating the effectiveness of the rewriting steps. Some data still flagged as contaminated after the first rewrite maintain high accuracy but are reclassified as uncontaminated after multiple rewrites, leading to a significant reduction in contamination rate. This analysis confirms that iterative rewrites effectively transform contaminated data into reliable, uncontaminated data, ensuring dataset accuracy and integrity.

\section{Conclusion}
This paper explores eliciting truthful answers from a language model by addressing the impact of data contamination on model evaluations. We propose an inference-time decontamination method involving detection and iterative rewriting of contaminated data, leading to more accurate model performance assessments. Experiments on GSM8K and MMLU benchmarks suggest that our method can mitigate contamination effects, resulting in more reliable evaluation results.

Our framework's detection, rewrite, and assurance stages allow for consistent and fair assessments without needing entirely new datasets. The reduction in contamination's impact highlights the promise of our approach in providing a realistic view of model capabilities.

We believe this work lays a foundation for future research in improving language model evaluations. Further exploration of advanced detection and rewriting techniques will continue to enhance the reliability and fairness of these assessments.

\section*{Limitations}
\paragraph{Limited evaluation criteria for Real Models:} The correction magnitude for real models is not substantial. However, since the specific contamination relationship between a model and the benchmark is still unknown, it is impossible to provide an effective evaluation without a reliable detecting method.
    
\paragraph{Effectiveness in the Worst-case Scenarios:} For models that have intentionally trained on a large amount of in-domain data to improve performance, or even followed our revival steps for extensive rewrites, our current two types of rewrite methods have limited effectiveness. Nonetheless, due to the detect stage in our setting, we can distinguish these models from those with no contamination. We also look forward to exploring more efficient rewrite methods in the future.
    
\paragraph{Evaluation Biases} The different contamination levels across models lead to variations in the final evaluation questions. Additionally, using model-automated rewriting to reduce costs may introduce biases. We discuss these issues here. Firstly, as shown in our Quality Checks experiments, we ensure that each rewrite maintains consistent difficulty and tests the same knowledge and skill points. Secondly, the bias introduced by model generation is relatively small compared to the false score increase from models memorizing original questions. This randomness affects all models. Furthermore, we use cached rewritten datasets to reduce costs and randomness, thereby increasing relative comparability between models.

However, the above points regarding the detecting methods and rewriting methods are not inherent limitations of our setting. They can be mitigated by using better rewriting and detecting methods in the future.

\bibliography{custom}

\appendix
\label{sec:appendix}
\section{Experiment on Hyper Parameter}
\label{sec:appendix-hper}
\begin{figure}[t]
\centering
\subfigure{
\includegraphics[width=0.3\textwidth]{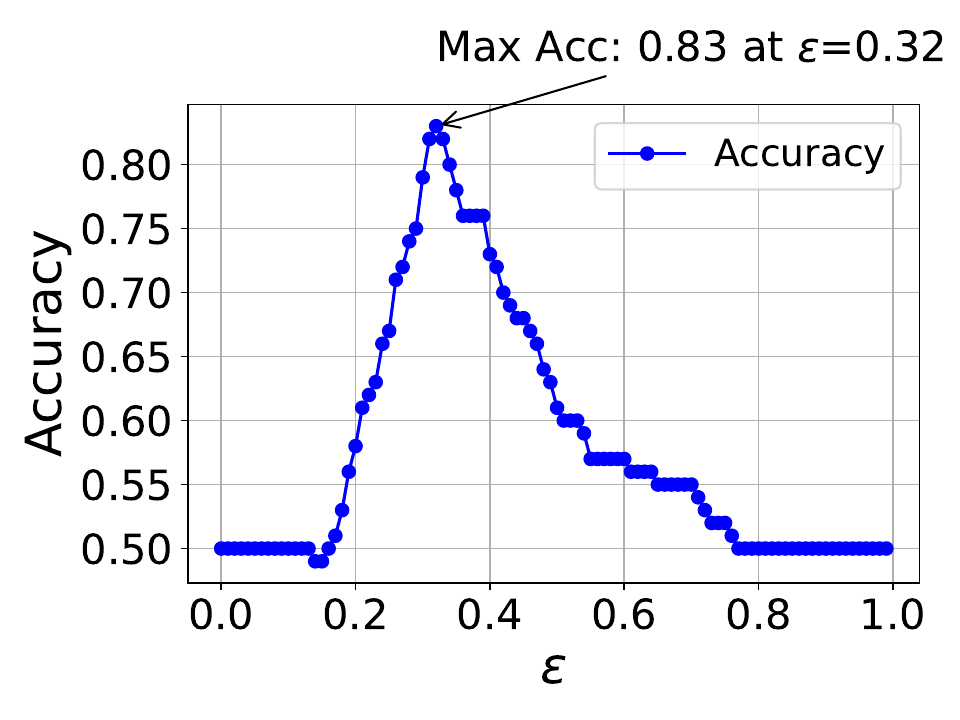}
}
\subfigure{
\includegraphics[width=0.3\textwidth]{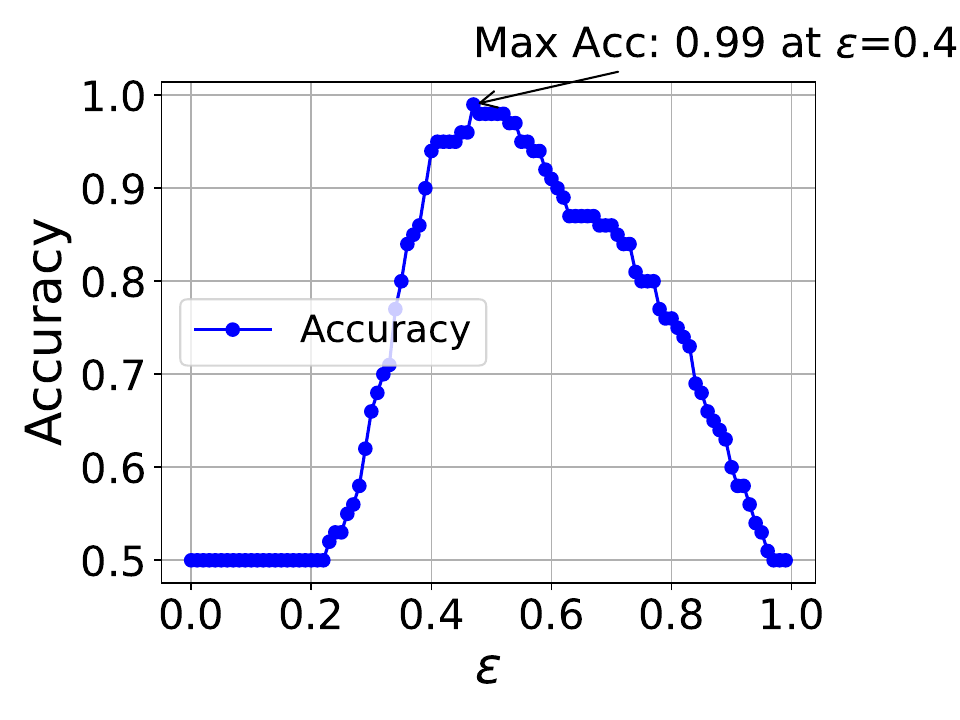}
}
\subfigure{
\includegraphics[width=0.3\textwidth]{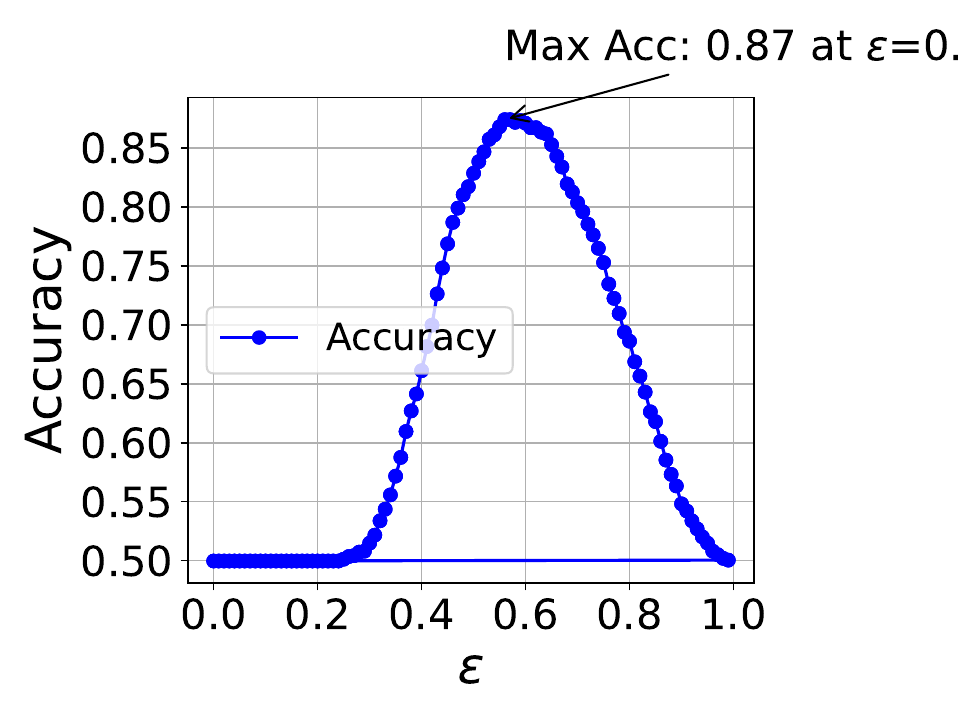}
}
\vspace{-5pt}
\caption{Hyper Parameter Search Experimen about \( \epsilon \) on GSM8K.\vspace{-16pt}}
\label{fig:hper-gsm8k}
  
\end{figure}

\begin{figure}[t]
\centering
\subfigure{
\includegraphics[width=0.3\textwidth]{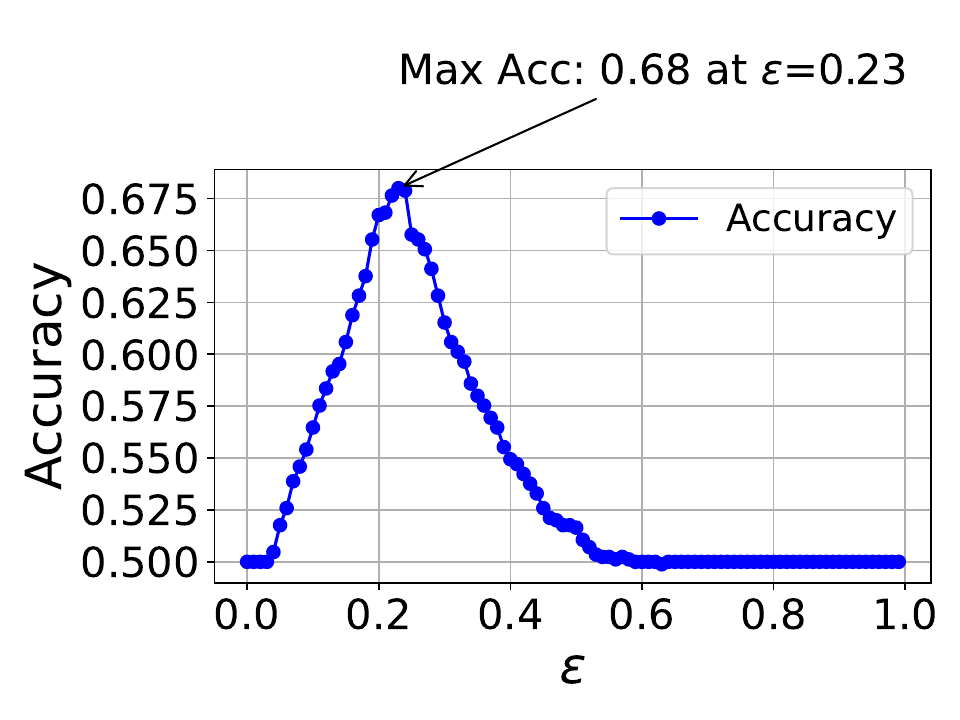}
}
\subfigure{
\includegraphics[width=0.3\textwidth]{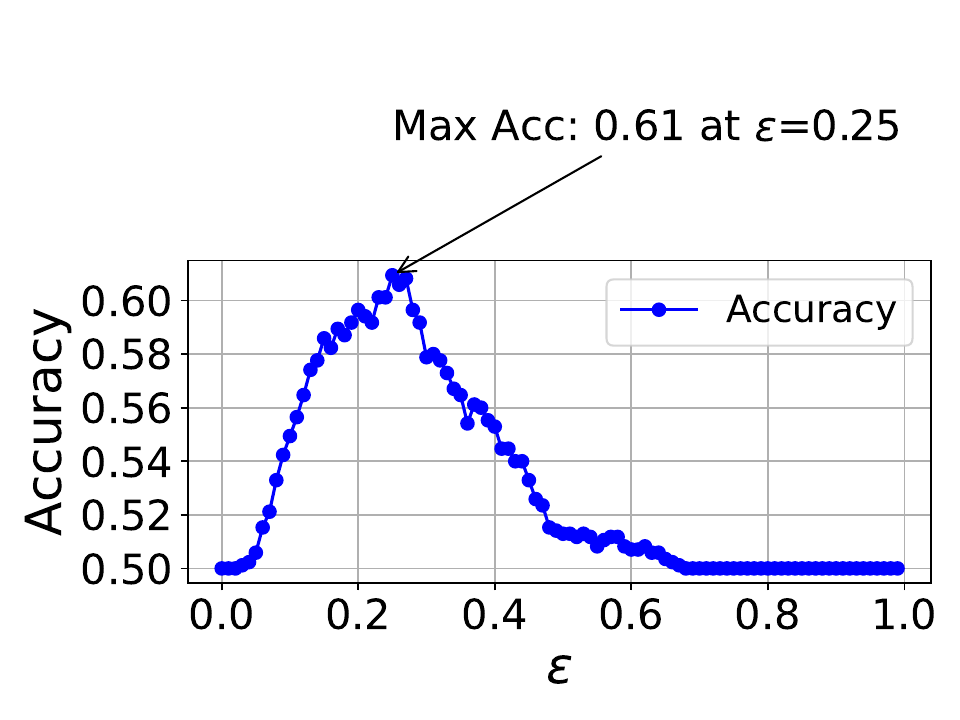}
}
\subfigure{
\includegraphics[width=0.3\textwidth]{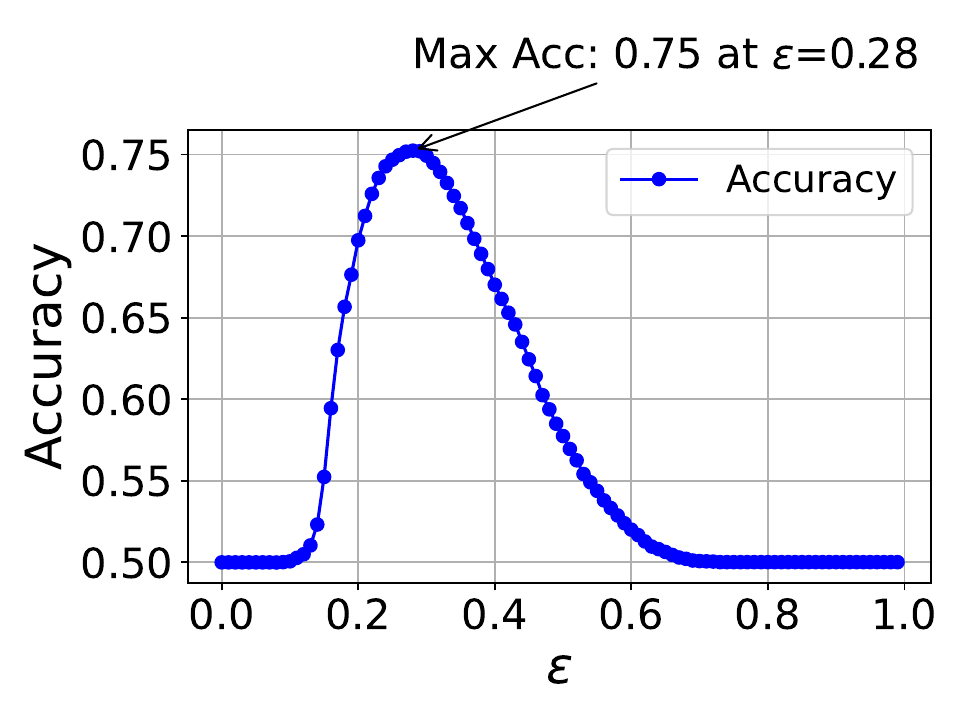}
}
\vspace{-5pt}
\caption{Hyper Parameter Search Experimen about \( \epsilon \) on MMLU.\vspace{-16pt}}
\label{fig:hper-mmlu}
  
\end{figure}

\begin{figure}[t]
\centering
\subfigure{
\includegraphics[width=0.3\textwidth]{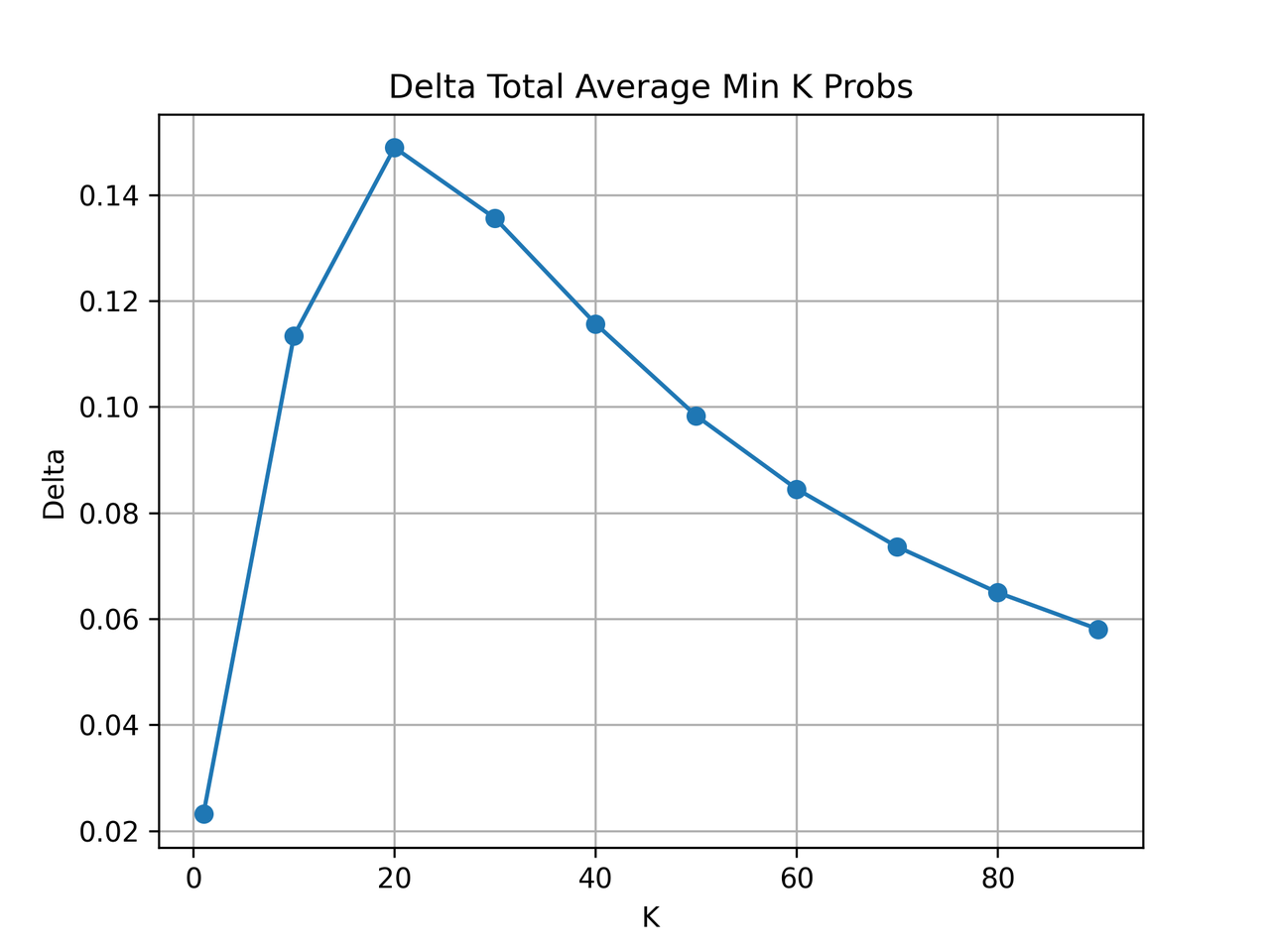}
}
\subfigure{
\includegraphics[width=0.3\textwidth]{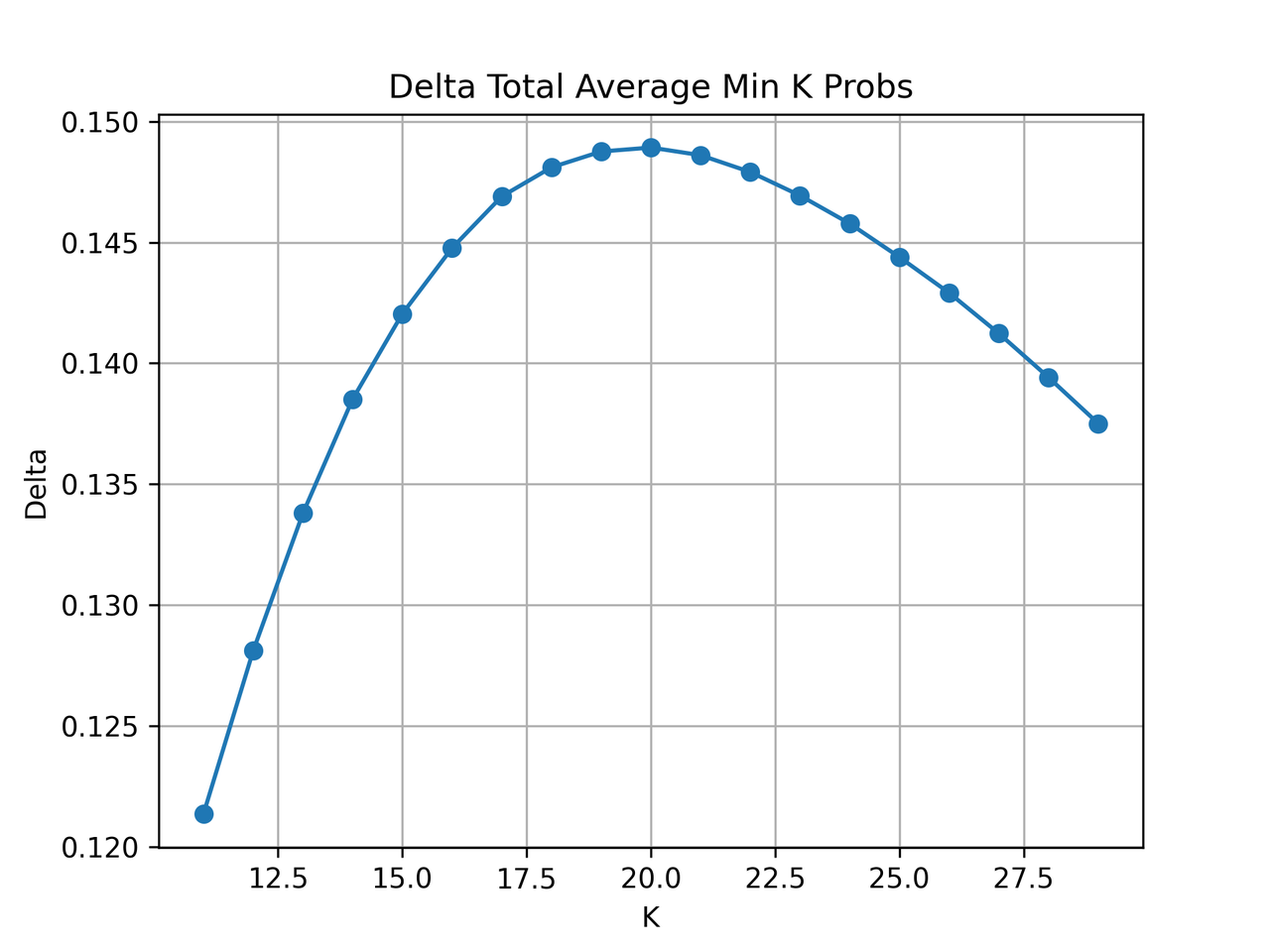}
}
\vspace{-5pt}
\caption{Hyper Parameter Search Experimen about K\vspace{-16pt}}
\label{fig:hper-k}
  
\end{figure}

In the implementation of the detector, we need to determine two hyperparameters: \( K \) and \( \epsilon \).
The selection criterion for \( K \) is the difference in MinKprob before and after rewriting, aiming for sensitivity to the impact of rewriting. For \( \epsilon \), Shi originally required obtaining some seen and unseen data corresponding to the model. However, this is cumbersome and challenging to find seen and unseen data with a similar domain under the setting of any black-box model. Therefore, our approach is to artificially construct a part of the seen set through training and approximate an unseen set through multiple rewrites. Specifically, we randomly sample from the test set and perform several rewrites. Then, we uniformly divide the rewritten test set \( D \) into seen and unseen sets. By training, we artificially expose the seen data. We then iterate over the threshold \( \epsilon \) and perform binary classification on each data point in the rewritten \( D \) to determine whether it has been seen, selecting the threshold \( \epsilon \) that yields the most accurate classification under a uniform distribution based on accuracy.
\paragraph{Experiment on \( \epsilon \)}
By exhaustively searching for the highest classification accuracy on the constructed seen and unseen sets, We determined \( \epsilon \) as shown in Figure~\ref{fig:hper-gsm8k},\ref{fig:hper-mmlu}. 
\paragraph{Experiment on K}
By exhaustively searching for the maximum difference in MinKprob before and after rewriting, we determined \( K = 20 \) as shown in Figure~\ref{fig:hper-k}.

\section{Rewrite details}
\label{sec:rewrite-details}
\subsection{Prompt}

In this section, we describe the process used to rephrase the stems of fill-in-the-blank math questions using GPT-4. 

For knowledge-related benchmarks MMLU, we keep the knowledge points tested by the original sample unchanged and rewrite the phrasing of the questions.

For math benchmarks related to model reasoning abilities GSM8K, we maintain the specific numbers and calculations involved in the original data unchanged, but rewrite the background of the questions.

The prompts used to guide the rephrasing process are designed to maintain the integrity and difficulty of the original questions while introducing diversity in the context and entities. The rephrasing involves the following guidelines:
\paragraph{Mathematical Reasoning Problems}
\begin{enumerate}
    \item The rephrased questions should not be a direct synonym replacement but can involve changing scenes and entities, such as replacing ``eggs'' with ``candies,'' as long as the numbers, operational logic, and final answers remain unchanged.
    \item The main goal is to retain the precision needed for the correct fill-in response.
    \item Ensure the semantics of the question stems are consistent before and after the rewrite.
    \item The revised version should not introduce biases or clues that could unfairly simplify the question.
    \item The rephrased question should be clear, concise, and maintain the original context and complexity.
    \item The rewritten stem should facilitate a step-by-step problem-solving process without affecting the expected mathematical solution.
    \item Changes to names are allowed as long as they do not confuse the identities of the characters involved.
\end{enumerate}
This structured prompt ensures clarity in the rephrasing process and maintains the quality and difficulty level of the math questions.

We show the prompt in Table~\ref{tab:prompt_math}.

\begin{figure}[t]
\centering
\subfigure[Rewritten example for Mathematical Reasoning Problems]{
\includegraphics[width=0.5\textwidth]{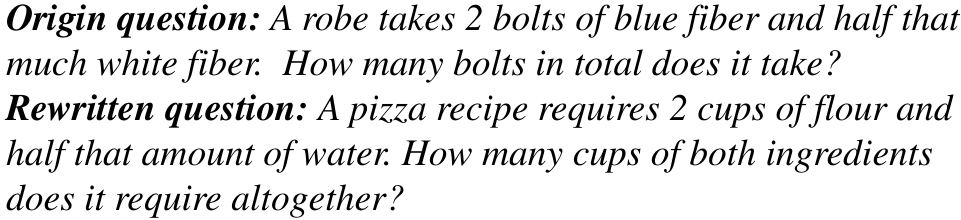}
}
\subfigure[Rewritten examples]{
\includegraphics[width=0.5\textwidth]{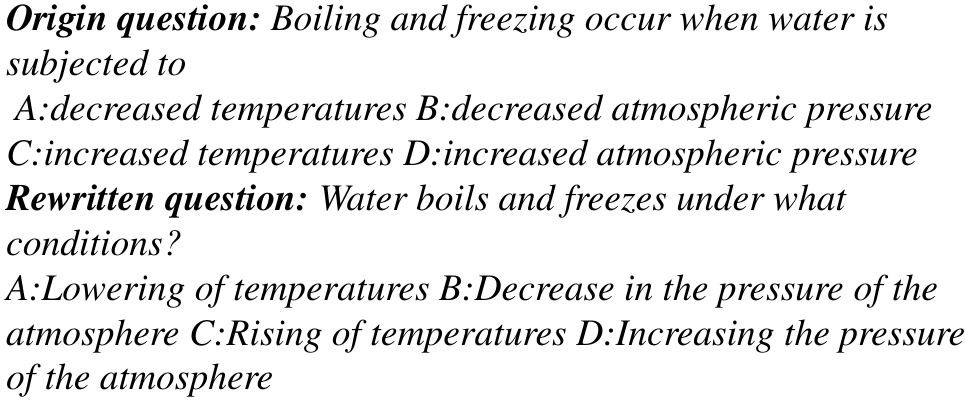}
}

\vspace{-5pt}
\caption{Hyper Parameter Search Experimen about \( \epsilon \) on GSM8K.\vspace{-16pt}}
\label{fig:written-example}
  
\end{figure}

\paragraph{Knowledge-based Problem}
\begin{enumerate}
\item The revised questions should maintain the original meaning and accuracy without any bias or hinting at the correct answer.
\item Ensure that the difficulty of understanding and solving the problem remains consistent before and after rewriting.
\item Use diverse expressions to avoid repetition but do not intentionally use uncommon words.
\item Do not alter mathematical expressions.
\end{enumerate}
This structured prompt ensures clarity in the rephrasing process and maintains the quality and difficulty level of the knowledge-based questions.

We show the prompt in Table~\ref{tab:prompt_knowledge}.

\subsection{Written Examples}
We randomly sampled some rewritten examples of the two benchmarks.
Examples of single-round rewrites can be found in Figure~\ref{fig:written-example}. And examples of multi-round rewrites can be found in the Table~\ref{tab:gsm8k-example},~\ref{tab:mmlu-example}.

More examples can be seen in the three-round rewritten data on the entire dataset we released.

\begin{table*}[h]
    \centering
    \small
    \begin{tabular}{p{14cm}}

\toprule
Your task involves revising the stems of fill-in-the-blank math questions. For added diversity, this rewrite is not a direct synonym replacement; you can change scenes and entities—like replacing eggs with candies—as long as the numbers, operational logic, and final answers remain unchanged, without altering the difficulty level. The primary goal is to rephrase each question's stem—the main question or statement—in a way that retains the precision needed for the correct fill-in response. Ensure that the semantics of the question stems are consistent before and after the rewrite. The revised version should not introduce biases or clues that could unfairly simplify the question. It should be clear, concise, and maintain the original context and complexity. Furthermore, the rewritten stem should facilitate a step-by-step problem-solving process without affecting the expected mathematical solution, allowing changes to names as long as they do not confuse the identities of the characters involved.\\
\\
Here are two examples for better understanding. Follow them and answer in json format:\\
\\
Input:\\
Original\_Question\_Stem:\\
"Natalia sold clips to 48 of her friends in April, and then she sold half as many clips in May. How many clips did Natalia sell altogether in April and May?"\\
Answer:\\
"Natalia sold 48/2 = $<<$48/2=24$>>$24 clips in May. Natalia sold 48+24 = $<<$48+24=72$>>$72 clips altogether in April and May. \#\#\#\# 72"\\
Output:\\
$\lbrack$\\
    \{\\
        "Rephrased\_Question\_Stem": "In March, Marco gathered 48 seashells at the beach, and in April, he collected half as many. How many seashells did Marco collect in total during March and April?",\\
        "Rephrased\_Answer": "Marco collected 48/2 = 24 seashells in May. Marco collected 48 + 24 = 72 seashells altogether in April and May. \#\#\#\# 72",\\
    \}\\
$\rbrack$\\
Input:\\
Original\_Question\_Stem:\\
"Weng earns \$12 an hour for babysitting. Yesterday, she just did 50 minutes of babysitting. How much did she earn?"\\
Answer:\\
"Weng earns 12/60 = \$<<12/60=0.2>>0.2 per minute. Working 50 minutes, she earned 0.2 x 50 = \$<<0.2*50=10>>10. \#\#\#\# 10"\\
Output:\\
$\lbrack$\\
    \{\\
        "Rephrased\_Question\_Stem": "Each day, Kevin’s bees produce 16 tablespoons of honey. He uses three tablespoons to sweeten his morning tea and four to make energy bars for his hiking group. He sells the leftover honey at the local co-op for \$2 per tablespoon. How much money does Kevin earn daily from selling honey at the co-op?",\\
        "Rephrased\_Answer": "Kevin sells 16 - 3 - 4 = <<16-3-4=9>>9 tablespoons of honey a day. He earns 9 * 2 = \$<<9*2=18>>18 every day at the local co-op. \#\#\#\# 18",\\
    \}\\
$\rbrack$
\\
Start with this question and apply the instructions above:\\
Your Question Stem to Rephrase:\\

\bottomrule
    \end{tabular}
\caption{
    The instruction for revising math question stems.
    }
    \label{tab:prompt_math}

\end{table*}

\begin{table*}[h]
    \centering
    \small
    \begin{tabular}{p{14cm}}

\toprule
Revise the multiple-choice question and options to keep the meaning and accuracy without any bias or hinting at the correct answer.Ensure that the difficulty of understanding and solving the problem remains consistent before and after rewriting. Use diverse expressions to avoid repetition.But do not intentionally use uncommon words to avoid repetition and do not alter mathematical expressions. Follow the example and answer in JSON format:\\
\\
Input:\\
Original\_Question\_Stem: "During what historical period did the Renaissance take place?"\\
Original\_Options: "(A)The Late Middle Ages (B)The Classical Antiquity (C)The Enlightenment (D)The Industrial Revolution"\\
Output:\\
$\lbrack$\\
    \{\\
        "Rephrased\_Question\_and\_Options": \{\\
            "question": "In which historical era did the Renaissance occur?",\\
            "A": "The historic era just before the Renaissance",\\
            "B": "The period marking the transition from the Middle Ages",\\
            "C": "The historic era synonymous with the Age of Reason",\\
            "D": "The time period characterized by rapid industrialization"\\
        \}\\
    \}\\
$\rbrack$\\
\\
Input:\\
Original\_Question\_Stem: [[Original\_Question\_Stem]]\\
Original\_Options: [[Original\_Options]]\\
Output:\\

\bottomrule
    \end{tabular}
\caption{
    The instruction for revising multiple-choice questions and options.
    }
    \label{tab:prompt_knowledge}
\end{table*}

\begin{table*}[h]
    \centering
    \small
    \begin{tabular}{p{14cm}}
\toprule

\textbf{Original Question:} \\
"Every day, Wendi feeds each of her chickens three cups of mixed chicken feed, containing seeds, mealworms and vegetables to help keep them healthy. She gives the chickens their feed in three separate meals. In the morning, she gives her flock of chickens 15 cups of feed. In the afternoon, she gives her chickens another 25 cups of feed. How many cups of feed does she need to give her chickens in the final meal of the day if the size of Wendi's flock is 20 chickens?" \\
\midrule

\textbf{Original Answer:} \\
"If each chicken eats 3 cups of feed per day, then for 20 chickens they would need 3*20=60 cups of feed per day. If she feeds the flock 15 cups of feed in the morning, and 25 cups in the afternoon, then the final meal would require 60-15-25=20 cups of chicken feed. \#\#\#\# 20" \\
\midrule

\textbf{Rephrased Question Stem (Step 1):} \\
"Every day, Carla prepares 3 liters of homemade insect food, full of protein and vitamins, for her butterflies. She divides the feed for her swarm of butterflies into three meals. In the morning, her butterfly population consumes 15 liters of the feed. In the afternoon, they consume another 25 liters. How many liters of insect food does Carla need to prepare for the final meal if she has 20 butterflies?" \\
\midrule

\textbf{Rephrased Answer (Step 1):} \\
"Each butterfly consumes 3 liters of feed daily, meaning for 20 butterflies Carla would need 3*20 = 60 liters of feed in a day. If they consume 15 liters in the morning, and 25 liters in the afternoon, then the final meal requires 60-15-25 = 20 liters of insect food. \#\#\#\# 20" \\
\midrule

\textbf{Rephrased Question Stem (Step 2):} \\
"At home, Julia bakes 3 chew sticks each day for her 20 parrots. She breaks down these sticks into three increments throughout the day. In the morning alone, her parrots munch down on 15 of these sticks, and they gnaw on another 25 in the middle of the day. How many chew sticks does Julia need to bake for the final session of the day?" \\
\midrule

\textbf{Rephrased Answer (Step 2):} \\
"Each parrot requires 3 chew sticks a day, so for her 20 parrots, Julia needs to bake 3*20 = 60 chew sticks every day. If they consume 15 in the morning and 25 in the afternoon, then the final munching session would require 60-15-25 = 20 chew sticks. \#\#\#\# 20" \\
\midrule

\textbf{Rephrased Question Stem (Step 3):} \\
"Simon has a habit of preparing 3 kilograms of bird food, filled with nutritious grains, for his parakeets every day. To cater to their feeding habits, he splits the food into three meals. In the morning, his flock of parakeets consumes 15 kilograms of bird food. Later in the afternoon, they consume another 25 kilograms. Given that he has 20 parakeets, how many kilograms of bird food does Simon need to portion out for their final feed of the day?" \\
\midrule

\textbf{Rephrased Answer (Step 3):} \\
"Each parakeet consumes 3 kilograms of bird food daily, which means that Simon would need to prepare 3*20 = 60 kilograms of bird food in a day for 20 parakeets. If 15 kilograms are consumed in the morning, and 25 kilograms in the afternoon, then Simon will need to prepare an additional 60-15-25 = 20 kilograms of bird food for the final meal. \#\#\#\# 20" \\
\bottomrule
    \end{tabular}
\caption{
    Written examples for gsm8k.
}
\label{tab:gsm8k-example}
\end{table*}

\begin{table*}[h]
    \centering
    \small
    \begin{tabular}{p{14cm}}
\toprule

\textbf{Original Question:} \\
"A sixth-grade teacher is concerned because Kerry, a student in class, has been hostile to classmates. Which of the following teacher strategies is most likely to encourage Kerry to be more cooperative with classmates?" \\
\midrule

\textbf{Original Options:} \\
A: "Preventing Kerry from participating in play or recess activities as a consequence of hostile behavior" \\
B: "Having Kerry memorize rules of behavior and write examples of how they would apply in the classroom" \\
C: "Withholding attention or approval from Kerry in response to hostile behavior" \\
D: "Implementing social skills training to teach Kerry appropriate replacement behaviors for hostile behaviors" \\
\midrule

\textbf{Rephrased Question and Options (Step 1):} \\
\textbf{Question:} "A student named Kerry in the sixth grade has been acting unfriendly towards peers. Which method could the teacher possibly use to promote more amicable relationships among Kerry and her classmates?" \\
A: "Prohibit Kerry from joining in leisure or break-time activities as a repercussion for unfriendly conduct" \\
B: "Instruct Kerry to learn behavior norms by heart, and to pen down how they can be enforced within the classroom" \\
C: "Deny Kerry attention or appreciation when hostile conduct is exhibited" \\
D: "Incorporate social abilities development to instruct Kerry on acceptable alternate behaviors to her antagonistic actions" \\
\midrule

\textbf{Rephrased Question and Options (Step 2):} \\
\textbf{Question:} "What strategy could a teacher employ to encourage Kerry, a sixth-grader, to cultivate better relationships with her peers, given her recent hostile behavior?" \\
A: "Exclude Kerry from recreational breaks as punishment for her negative conduct" \\
B: "Request Kerry to internalize good behavioral practices and document their implementation in class" \\
C: "Avoid acknowledging or praising Kerry if she displays aggressive behavior" \\
D: "Use techniques promoting the development of interpersonal skills to address and modify Kerry's non-cooperative behavior" \\
\midrule

\textbf{Rephrased Question and Options (Step 3):} \\
\textbf{Question:} "What approach could an educator utilize to motivate Kerry, a student in grade six, to improve her associations with classmates, considering her lately antagonistic behavior?" \\
A: "Bar Kerry from leisure intervals as retribution for her adverse behavior" \\
B: "Ask Kerry to absorb constructive behavioral norms and record their enactment in the classroom" \\
C: "Overlook or restrain from complimenting Kerry if she exhibits hostility" \\
D: "Apply strategies fostering the enhancement of social abilities to handle and transform Kerry's uncooperative conduct" \\
\bottomrule
    \end{tabular}
\caption{
    Written examples for MMLU:high school psychology.
}
\label{tab:mmlu-example}
\end{table*}

\end{document}